\documentclass[twoside]{article}
\usepackage[utf8]{inputenc}
\usepackage{amsmath}
\usepackage{amsfonts}
\usepackage{amssymb}
\usepackage{graphicx}
\usepackage{color}
\usepackage[normalem]{ulem}
\usepackage{algorithm,algorithmic,multirow}
\usepackage[left=2cm,right=2cm,top=2cm,bottom=2cm]{geometry}
\newenvironment{Proof}{\noindent{\sc Proof.}}{\qed}
\newtheorem{theorem}{Theorem}[section]
\newtheorem{lemma}{Lemma}[section]
\newtheorem{cor}{Corollary}[section]
\newtheorem{rem}{Remark}[section]
\newtheorem{definition}{Definition}[section]
\newtheorem{prop}{Proposition}[section]

\newcommand{\qed}{\hfill$\Box$\par\medskip}
\renewcommand{\theequation}{\arabic{section}.\arabic{equation}}
\def\bhag#1{\noindent
\setcounter{equation}{0}
\section{#1}
}

\def\HH{{\mathbb H}}

\def\RR{{\mathbb R}}
\def\CC{{\mathbb C}}

\def\PPI{{{\rm I}\kern-1pt\Pi}}
\def\SS{{\mathbb S}}

\def\b #1;{{\bf #1}}
\def\x{{\bf x}}

\def\y{{\bf y}}

\def\O{{\cal O}}

\def\C{{\mathcal C}}

\def\esssup{\mathop{\hbox{{\rm ess sup}}}}

\def\be{\begin{equation}}
\def\ee{\end{equation}}
\def\bea{\begin{eqnarray}}
\def\eea{\end{eqnarray}}
\def\eref#1{(\ref{#1})}
\def\disp{\displaystyle}

\def\donchitre#1#2{\vskip 6.5cm\noindent
\parbox[t]{1in}{\special{eps:#1.eps x=6.5cm y=5.5cm}}
\hbox to 7cm{}\parbox[t]{0.0cm}{\special{eps:#2.eps x=6.5cm y=5.5cm}}}

\def\tn{|\!|\!|}

\def\bs#1{{\boldsymbol{#1}}}

\def\supp{\mathsf{supp\ }}

\title{Function approximation with   zonal function  networks with activation functions analogous to the rectified linear unit functions }
\author{
 H.~N.~Mhaskar\thanks{
Institute of Mathematical Sciences, Claremont Graduate University, Claremont, CA 91711. The research of this author is supported in part by ARO Grant W911NF-15-1-0385, and  by the Office of the Director of National Intelligence (ODNI), Intelligence Advanced Research Projects Activity (IARPA), via 2018-18032000002.
\textsf{email:} hrushikesh.mhaskar@cgu.edu} 
 }
 \date{July 8, 2018}
\begin{document}
\maketitle
\begin{abstract}
A zonal function (ZF) network on the $q$ dimensional sphere $\SS^q$ is a network of the form $\x\mapsto \sum_{k=1}^n a_k\phi(\x\cdot\x_k)$ where $\phi :[-1,1]\to\RR$ is the activation function,  $\x_k\in\SS^q$ are the centers, and $a_k\in\RR$. 
While the approximation properties of such networks are well studied in the context of positive definite activation functions, recent interest in deep and shallow networks motivate the study of activation functions of the form $\phi(t)=|t|$, which are not positive definite. 
In this paper, we define an appropriate smoothess class and establish approximation properties of such networks for functions in this class. 
The centers can be chosen independently of the target function, and the coefficients are linear combinations of the training data. The constructions preserve rotational symmetries.
\end{abstract}

\bhag{Introduction}\label{intsect}

A neural network is a function of the form $\x\mapsto
\sum_{k=1}^n a_k\phi(\x\cdot\x_k+b_k)$, $\x, \x_k\in\RR^q$, $a_k\in\RR$, and $\phi: \RR^q\to \RR$ is an \emph{activation function}. 
An attractive property of these networks is that unlike their predecessors, perceptrons \cite{minsky1988perceptrons}, they have the \emph{universal approximation property}: any continuous function on  any compact subset of any Euclidean space $\RR^q$ 
can be approximated arbitrarily well by neural networks with very minimal conditions on the activation function. 
Clearly, neural networks are also important from the point of view of computation in that they can be evaluated in parallel. There are many journals devoted to the topic of neural networks and their properties.

Approximation of multivariate functions by neural networks is a very old topic  with some of the first papers dating back to the late 1980s e.g., \cite{irie1988,  cybenko1989, funahashi1989, stinchcombe1989}.  
Some very general conditions on the activation function that permits the universal approximation property are given in \cite{mhasmich, leshnolinpinkus}.
An important question in this theory is to estimate the number of nonlinear units in the network required to approximate a given target function up to a given accuracy. 
Clearly, as with all similar questions in classical approximation theory, the answer to this question depends upon how one  measures the ``smoothness'' of the target function. Thus, various dimension independent bounds are given in \cite{barron1993, dimindbd, tractable}. 
When the smoothness is measured by the number of derivatives, the optimal estimate consistent with that in the classical theory of polynomial approximation   is given in \cite{optneur}. A survey of these ideas can be found in \cite{pinkus1999approximation, indiapap}.

 The interest in this subject is renewed recently with the advent of deep networks.
 Deep neural networks especially of the convolutional type (DCNNs) have
started a revolution in the field of artificial intelligence and
machine learning, triggering a large number of commercial ventures and
practical applications. Most deep learning references
these days start with Hinton's backpropagation and with LeCun's
convolutional networks (see for a nice review
\cite{lecun2015deep}). We have started an investigation of approximation properties of deep networks in \cite{dingxuanpap}, where it is shown that one reason why deep networks are so effective is that they allow one to utilize any compositional structure in the target function. 
A popular activation function in the theory of deep networks is the ReLU (Rectified linear unit) function $t\mapsto t_+=\max(t, 0)$. 
One reason for this popularity is the relative ease with which one can formulate training algorithms for multiple layers with this activation function \cite{choromanska2015loss}. 
It is pointed out in \cite{bach2014, dingxuanpap} that the problem of studying
function approximation on a Euclidean space by shallow networks evaluating the ReLU activation function is equivalent to the problem of studying approximation of even functions on the sphere by ZF networks evaluating this activation function.  
Moreover, results about shallow networks can be lifted to those about deep networks using what we have called \emph{good propagation property}.

Many problems in geophysics also lead to the question of approximation of functions on the sphere. Therefore, 
 we studied in \cite{mnw1, zfquadpap} approximation on the sphere by neural networks. Unlike the setting of the Euclidean space, translation is not possible anymore, and the neural networks take the form of what we have called \emph{Zonal Function (ZF) networks}. 
 A zonal function (ZF) network on the $q$ dimensional sphere $\SS^q$ is a network of the form $\x\mapsto \sum_{k=1}^n a_k\phi(\x\cdot\x_k)$ where $\phi :[-1,1]\to\RR$ is the activation function,  $\x_k\in\SS^q$ are the centers, and $a_k\in\RR$. 
 In general, this theory is applicable only when the activation function $\phi$ is positive definite (in addition to some other conditions).

The purpose of this paper is to develop an estimate on the degree of approximation by ZF networks where the activation functions include the  ReLU function. 
Since $|t|=t_+ + (-t)_+$ and $t_+=(t+|t|)/2$, approximation by networks with an activation function given by $t\mapsto |t|$ is equivalent (up to a doubling of the count of nonlinearities or alternately, a linear term) to that by networks with the activation function $t\mapsto t_+$. 
The mathematical details are  more pleasant with the activation function $t\mapsto |t|$. 
These details generalize effortlessly to the case of the activation functions $\phi_\gamma(t)=|t|^{2\gamma+1}$, where $2\gamma+1$ is not an even integer.
The main challenge here is that none of these functions, restricted to the sphere, is  positive definite. Thus, the results in \cite{zfquadpap} are not applicable. 

  Our proofs are constructive, and the constructions  have several interesting properties. 
\begin{enumerate}
\item The constructions do not require any machine learning in the classical sense. Given a data of the form $(\xi, f(\xi))$, for $\xi$ in some finite subset of $\SS^q$, we give explicit formulas for the approximating networks (cf. \eref{zfnetexplicit}).
\item The centers of the networks are chosen independently of the target function or the training data (cf. \eref{quadratureexplicit} and \eref{zfnetexplicit}).
\item We do not assume any priors on the target function; the constructions are the same for every continuous function on $\SS^q$, with estimates in terms of its smoothness adjusting automatically. 
\item The networks are invariant under rotations of $\RR^{q+1}$.  (cf. Remark~\ref{sphrotinvariancerem})
\item The coefficients are linear functionals on the target function $f$, and if the available data consists of spherical harmonic coefficients of a sufficiently ``smooth'' $f$, then the sum of absolute values of these coefficients is bounded independently of the size of the network/amount of available data (cf. \eref{sphcoefftheory}).
\end{enumerate}

In Section~\ref{relationsect}, we discuss the relationship of our paper with a few other recent relevant papers.  After reviewing some preliminary facts about analysis on the sphere in Section~\ref{perlimsect}, we state our main result and algorithm in Section~\ref{mainsphapproxsect}. The proof of the result requires several preparatory technical details which are reviewed in Section~\ref{prepsecct}. The proof itself  is given in Section~\ref{sphapproxsect}. In the Appendix, we derive an expansion of the function $\phi_\gamma(t)=|t|^{2\gamma+1}$ in terms of certain ultraspherical polynomials.

\bhag{Relation to other work}\label{relationsect}
\begin{enumerate}
\item In \cite{zfquadpap}, we have studied approximation by ZF networks with a positive definite  activation function  (i.e., one that admits an expansion of the form 
$$
\sum_{\ell=0}^\infty a_\ell p_\ell^{(q/2-1,q/2-1)}(1)p_\ell^{(q/2-1,q/2-1)}(\circ),
$$
 where $a_\ell$'s are all positive), and $a_\ell\sim\ell^{-\beta}$ for some $\beta>0$ in a technical sense described in this paper. (See Section~\ref{sphprelimsect} below for notation.)
These results are not applicable in the current context, since the ReLU function does not satisfy these conditions.
\item
Our current paper depends heavily on the special function connections between ultraspherical and Jacobi polynomials, in addition to dealing with activation functions that are not positive definite. These ideas are not applicable in the case of approximation on a general manifold.
\item In \cite{bach2014}, the authors have studied approximation by ReLU networks on the sphere. 
The degree of approximation in our paper is better than that in \cite{bach2014}. Also, unlike the proof in that paper, we give explicit constructions for our approximations.
\item In \cite{yarotsky2016error, schmidt2017nonparametric}, a deep network with ReLU activation function is shown to approximate a function on a Euclidean cube, if the target function satisfies some severe conditions. The error bounds are probabilistic in nature.
Together with the ideas in \cite{dingxuanpap}, our results in this paper give constructive proofs for deterministic bounds for approximation on the whole Euclidean space for both shallow and deep networks. Our results are applicable when the deep networks are described be a general directed acyclic graph rather than a tree.
\item In \cite{klusowski2016uniform}, the authors obtain estimates on the degree of approximation on a Euclidean cube by networks using the ReLU (or its square) as the activation function. The target function is assumed to satisfy an integral expression analogous to \eref{fundaidentity} below.  These bounds are ``dimension independent''. Similar bounds for approximation on the entire Euclidean space were obtained also in \cite{tractable} (and references therein) for a more general class of activation functions. 
The proofs of all these results are based on certain probabilistic estimates, and are not constructive. 

\end{enumerate}
\bhag{Preliminaries}\label{perlimsect}

The statement of our main result requires some notation and background preparation. In Section~\ref{sphprelimsect}, we develop the necessary notation. In Section~\ref{regmeasuresect}, we review the notion of Marcinkiewicz-Zygmund (MZ) quadrature measures on the sphere, which play a critical role in our construction.
In Section~\ref{smoothsect}, we define a class of activation functions 
of interest in this paper, including the ReLU functions, define the associated continuous networks and smoothness classes, as well as review the properties of 
certain operators essential in our constructions.
 
\subsection{Notation on the sphere}\label{sphprelimsect}

In the sequel, $\mu_q^*$ denotes the Lebesgue surface (Riemannian volume) measure of $\SS^q$. The surface area of $\SS^q$ is $\disp \omega_q= \frac{2\pi^{(q+1)/2}}{\Gamma((q+1)/2)}$. The geodesic distance between points $\x,\y\in\SS^q$ is given by $\rho(\x,\y)=\arccos(\x\cdot\y)$.
For $r>0$, a spherical cap with radius $r$ and center $\x_0\in\SS^q$ is defined by
$$
\mathbb{B}(\x_0,r)=\{\y\in\SS^q : \rho(\x_0,\y)\le r\}.
$$
We note that for any $\x_0\in\SS^q$ and $r>0$,
\be\label{ballmeasurecond}
\mu_q^*(\mathbb{B}(\x_0,r))=\omega_{q-1}\int_0^r \sin^{q-1} t dt \le \frac{\omega_{q-1}}{q} r^q.
\ee
In the sequel, the term measure will mean a (signed or positive), complete, sigma-finite, Borel measure on $\SS^q$. The total variation measure of $\nu$ will be denoted by $|\nu|$. If $\nu$ is a measure, $1\le p\le \infty$,  and $f: \SS^q\to\RR$ is  $\nu$-measurable, we write
$$
\|f\|_{\nu;p} :=\left\{\begin{array}{ll}
\disp\left\{\int_{\SS^q} |f(\x)|^pd|\nu|(\x)\right\}^{1/p}, & \mbox{if $1\le p <\infty$,}\\
|\nu|-\esssup_{\x\in \SS^q} |f(\x)|, & \mbox{if $p=\infty$.}
\end{array}\right. 
$$
The space of all $\nu$-measurable functions on $\SS^q$ such that $\|f\|_{\nu;p} <\infty$ will be
denoted by $L^p(\nu)$, with the usual convention that two functions are considered equal as
elements of this space if they are equal $|\nu|$-almost everywhere. We will omit the mention of $\nu$ if $\nu=\mu_q^*$, unless we feel a cause for any confusion on this account. For example, $L^p=L^p(\mu_q^*)$, $\|f\|_p=\|f\|_{\mu_q^*;p}$.
The symbol $C(\SS^q)$ denotes the class of all  continuous,  real valued functions on $\SS^q$, equipped with the norm $\|\circ\|_{\infty}$. 

For a real number $x\ge 0$, let $\Pi_x^q$ denote the class of all spherical polynomials (i.e., the restrictions to $\SS^q$ of polynomials in $q+1$ variables) of degree $\le x$. (The class $\Pi_x^q$ is the same as the class $\Pi_n^q$, where $n$ is the largest integer not exceeding $x$. However, our extension of the notation allows us, for example, to use the simpler notation $\Pi_{n/2}^q$ rather than the more cumbersome notation $\Pi_{\lfloor n/2\rfloor}^q$.)
For a fixed integer $\ell\ge 0$, the restriction to $\SS^q$ of a
homogeneous harmonic polynomial of exact degree $\ell$ is called a spherical
harmonic of degree $\ell$. Most of the following information is based
on \cite{mullerbk}, \cite[Section~IV.2]{steinweissbk}, and \cite[Chapter XI]{batemanvol2}, although we use a
different notation. The class of all spherical harmonics of degree
$\ell$ will be denoted by $\HH^q_\ell$. The
spaces $\HH^q_\ell$ are mutually orthogonal relative to
the inner product of $L^2$. For any integer $n\ge 0$, we have $\Pi^q_n = \bigoplus_{\ell=0
}^{n}\HH^q_\ell$. The dimension of $\HH^q_\ell$ is given by
\be\label{hkdim}
d\,^q_\ell := \dim \HH^q_\ell= \left\{
\begin{array}{cl}
\disp{\frac{2\ell+q-1}{\ell+q-1} {\ell+q-1 \choose \ell}}, & \mbox{if
}\ell\ge 1,\\[3ex]
1,& \mbox{if }\ell=0.
\end{array}
\right.
\ee
and that of $\Pi^q_n$ is $\sum_{\ell=0}^n d\,^q_\ell=d\,^{q+1}_n$. Furthermore,
$L^2=\mbox{\rm $L^2$--closure}\big\{\bigoplus_{\ell=0}^\infty \HH^q_\ell\big\}$.
Hence, if we choose an orthonormal basis
$\{Y_{\ell,k}\,:\,k=1,\ldots, d^q_\ell\}$ for each $\HH^q_\ell$, then
the set $\{Y_{\ell,k}\,: k=1,\ldots,
d\,^q_\ell\,\ell=0,1,\ldots\,\}$ is a complete orthonormal basis for $L^2$. 

One has the
well-known addition formula \cite{mullerbk} and \cite[Chapter XI, Theorem 4]{batemanvol2} connecting $Y_{\ell,k}$'s with Jacobi polynomials defined in \eref{rodrigues}:
\begin{equation}
\label{addformula}
 \sum_{k=1}^{d\,^q_\ell} Y_{\ell,k}(\x)\overline{Y_{\ell,k}(\y)} =
\omega_{q-1}^{-1} p_\ell^{(q/2-1,q/2-1)}(1)p_\ell^{(q/2-1,q/2-1)}(\x\cdot\y), \qquad
\ell=0,1,\cdots, 
\end{equation}
where each $p_\ell^{(q/2-1,q/2-1)}$ is a univariate polynomial of degree $\ell$ with positive leading coefficient, and one has the orthogonality relation
$$
\int_{-1}^1 p_\ell^{(q/2-1,q/2-1)}(t)p_k^{(q/2-1,q/2-1)}(t)(1-t^2)^{q/2-1}dt =\delta_{\ell,k}, \qquad \ell, k=0,1,\cdots.
$$
In particular, for $\x\in\SS^q$, $\ell=0,1,\cdots$,
\be\label{constchristoffel}
\sum_{k=1}^{d_\ell^q} |Y_{\ell,k}(\x)|^2=\omega_{q-1}^{-1} p_\ell^{(q/2-1,q/2-1)}(1)^2=\omega_q^{-1}\int_{\SS^q}\sum_{k=1}^{d_\ell^q} |Y_{\ell,k}(\y)|^2d\mu_q^*(\y)=d_\ell^q \omega_q^{-1}.
\ee
If $f\in L^1$, we define 
\be\label{sphcoeffdef}
\hat{f}(\ell,k)=\int_{\SS^q}f(\y)\overline{Y_{\ell,k}(\y)}d\mu_q^*(\y), \qquad k=1,\cdots,d_\ell^q, \ \ell=0,1,\cdots.
\ee
We note that if $f$ is an even function, then \eref{addformula} shows that $\hat{f}(2\ell+1, k)=0$ for $\ell=0,1,\cdots$. 

If $\phi :[-1,1]\to\RR$, we define formally
\be\label{unicoeffdef}
\hat{\phi}(\ell)=\frac{\omega_{q-1}}{p_\ell^{(q/2-1,q/2-1)}(1)}\int_{-1}^1 \phi(t)p_\ell^{(q/2-1,q/2-1)}(t)(1-t^2)^{(q-2)/2}dt,
\ee
so that we have the formal expansions
\be\label{uniformalexpansion}
\phi(\x\cdot\y)=\sum_{\ell=0}^\infty \hat{\phi}(\ell)\sum_{k=1}^{d_\ell^q}Y_{\ell,k}(\x)\overline{Y_{\ell,k}(\y)},
\ee
and
\be\label{sphereformalexpansion}
\int_{\SS^q}f(\y)\phi(\x\cdot\y)d\mu_q^*(\y)=\sum_{\ell=0}^\infty \hat{\phi}(\ell)\sum_{k=1}^{d_\ell^q}\hat{f}(\ell,k)Y_{\ell,k}(\x).
\ee

\noindent\textbf{Constant convention:}\\

In the sequel, the symbols $c, c_1,\cdots$ will denote generic positive constants depending only upon fixed parameters of the discussion such as $q$, the norm, the smoothness of the target functions, etc. Their values may be different at different occurrences even within the same formula. The notation $A\sim B$ means $c_1A\le B\le c_2A$.\\

\subsection{Regular measures on the sphere}\label{regmeasuresect}

The space of all signed (or positive), complete, sigma-finite, Borel measures on $\SS^q$ will be denoted by $\mathcal{M}$. If $\nu\in\mathcal{M}$, the set $\mathsf{supp}(\nu)$ is the set of all $x\in\SS^q$ with the property that $|\nu|(\mathbb{B}(x,r))>0$ for all $r>0$. It is easy to verify that $\mathsf{supp}(\nu)$ is a compact set.

\begin{definition}\label{absregularmeasuredef}
Let $d>0$. A  measure $\nu\in \mathcal{M}$  will be called \textbf{$ d$--regular} if
\begin{equation}\label{regulardef}
|\nu|(\mathbb{B}(\x,d))\le cd^q, \qquad \x\in\SS^q.
\end{equation}
The infimum of all constants $c$ which work in \eqref{regulardef} will be denoted by $|\!|\!|\nu|\!|\!|_{R,d}$, and the class of all  $d$--regular measures will be denoted by $\mathcal{R}_d$. 
\end{definition}
For example, $\mu_q^*$ itself is in  ${\cal R}_d$ with $|\!|\!|\mu_q^*|\!|\!|_{R,d}\le c$ for \emph{every} $d>0$ (see \eref{ballmeasurecond}).

The following proposition (cf. \cite[Proposition~5.6]{modlpmz}) reconciles different notions of regularity condition on measures defined in our papers.

\begin{prop}\label{mzequivprop}
Let $d\in (0,1]$, $\nu\in \mathcal{M}$. \\
{\rm (a)} If $\nu$ is $d$--regular, then for each $r>0$ and $\x\in\SS^q$,
\begin{equation}\label{regreconcile}
|\nu|(\mathbb{B}(\x,r))\le c|\!|\!|\nu|\!|\!|_{R,d}\ \mu^*(\mathbb{B}(\x,c(r+d)))\le  c_1|\!|\!|\nu|\!|\!|_{R,d}(r+d)^q.
\end{equation}
Conversely, if for some $A>0$, $|\nu|(\mathbb{B}(\x,r))\le A(r+d)^q$ or each $r>0$ and $\x\in\SS^q$, then $\nu$ is $d$--regular, and $|\!|\!|\nu|\!|\!|_{R,d}\le 2^q A$.\\
{\rm (b)} For each $\alpha>0$, 
\begin{equation}\label{regequiv}
|\!|\!|\nu|\!|\!|_{R,\alpha d}\le c_1(1+1/\alpha)^q |\!|\!|\nu|\!|\!|_{R,d}\le c_1^2(1+1/\alpha)^q(\alpha+1)^q|\!|\!|\nu|\!|\!|_{R,\alpha d},
\end{equation}
where $c_1$ is the constant appearing in \eqref{regreconcile}.\\
{\rm (c)} If $\nu$ is $d$--regular, then $\|P\|_{\nu;p}\le c_1\tn\nu\tn_{R,d}^{1/p}\|P\|_{\mu;p}$ for all $P\in \Pi_{1/d}$ and \textbf{all }
$1\le p<\infty$. Conversely, if for some $A>0$ and \textbf{some}
$1\le p<\infty$, $\|P\|_{\nu;p}\le A^{1/p}\|P\|_{\mu;p}$ for all $P\in \Pi_{1/d}$, then $\nu$ is $d$--regular, and $\tn\nu\tn_{R,d}\le c_2A$.
\end{prop}

In this paper, we are interested in quadrature formulas exact for integrating spherical polynomials, based on the training data; i.e., without assuming any specific location of the nodes involved in such a formula, in contrast to such well known formulas as the Driscoll-Healy or Clenshaw-Curtis formulas. 

\begin{definition}\label{quadmeasdef}
Let $n\ge 1$. A measure $\nu\in\mathcal{M}$ is called a \textbf{quadrature measure of order $n$} if
\begin{equation}\label{quadrature}
\int_{\SS^q} Pd\nu=\int_{\SS^q} Pd\mu^*, \qquad P\in\Pi_n^q.
\end{equation}
An \textbf{MZ (Marcinkiewicz-Zygmund) quadrature measure} of order $n$ is a quadrature measure $\nu$ of order $n$ for which $|\!|\!|\nu|\!|\!|_{R,1/n}<\infty$.
\end{definition}

If $\mathcal{C}\subset \SS^q$, we define the mesh norm $\delta(\mathcal{C})$ (also known as fill distance, covering radius, density content, etc.) and minimal separation $\eta(\mathcal{C})$ by
\begin{equation}\label{meshnormdef}
\delta(\mathcal{C})=\sup_{x\in\SS^q}\inf_{y\in \mathcal{C}}\rho(x,y), \qquad \eta(\mathcal{C})=\inf_{x, y\in \mathcal{C}, \ x\not=y}\rho(x,y).
\end{equation}

The first part of the following proposition (\cite[Lemma~5.3]{eignet}) gives an example of a regular measure apart from the surface measure itself. The second part (\cite[Theorem~5.7]{modlpmz}) asserts the existence of a regular measure supported on a given subset $\mathcal{C}$, that integrates polynomials of certain degree exactly, and the third part (\cite[Theorem~5.8]{modlpmz}) asserts that any positive quadrature formula necessarily defines a regular measure.
\begin{prop}\label{quadratureprop}
{\rm (a)} If $\mathcal{C}$ is finite, the measure that associates the mass $\eta(\mathcal{C})^q$ with each point of $\mathcal{C}$ is $\eta(\mathcal{C})$-regular, and $\tn\nu\tn_{R,\eta(\C)}\le c$. \\
{\rm (b)} There exist constants $c_1, c_2>0$ with the following property: If $\nu$ is a signed measure,\\ $\delta(\supp(\nu))< d<c_1$ and $0<n\le  c_2d^{-1}$, then there exists a simple function $W :\supp(\nu)\to [0,\infty)$, satisfying 
\be\label{diffpolyquad}
\int_{\SS^q} P(y)d\mu(y) =\int_{\SS^q} P(y)W(y)d|\nu|(y), \qquad P\in\Pi_n^q.
\ee
If $\nu$ is $d$--regular, then $W(y)\ge c\tn\nu\tn_{R,d}^{-1}$, $y\in\SS^q$.  \\
{\rm (c)} There exists a constant $c>0$ such that if $n\ge c$ and $\tau$ is a positive quadrature measure of order $2n$, then  for $1\le p<\infty$,
\be\label{wdnumzineq}
\|P\|_{\tau;p}\sim \|P\|_{\mu;p}, \qquad P\in \Pi_n^q,
\ee
where the constants involved may depend upon $p$ but not on $\tau$ or $n$. In particular, $\tau$ is $1/n$-regular, and $\tn\nu\tn_{R,1/n}\le c$.\\
\end{prop}

We  point out a consequence of Proposition~\ref{quadratureprop}. 
\begin{rem}\label{finitequadremark}
{\rm Let $\C$ be a finite subset of $\SS^q$. By removing close by points, we may obtain a subset $\tilde{C}\subseteq\C$ such that $\delta(\C)\sim\delta(\tilde{C})$ and $\eta(\tilde{C})\le 2\delta(\tilde{C})\le 4\eta(\tilde{C})$ (\cite[Proposition~2.1]{eignet}). We rename $\tilde{C}$ to be $\C=\{\x_1,\cdots,\x_M\}$. According to Proposition~\ref{quadratureprop}(a), the measure $\nu$ that associates the mass $\eta(\C)^q$ with each $\x_j$
is a $\eta(\C)$-regular measure with $\tn\nu\tn_{R, \eta(\C)}\le c$. Proposition~\ref{quadratureprop}(b) then asserts the existence of positive numbers $w_j$ such that
\be\label{finitequadrature}
\eta(\C)^q\sum_{j=1}^M w_jP(\x_j)= \int_{\SS^q} P(\x)d\mu_q^*(\x), \qquad P\in \Pi_{c\delta(\C)^{-1}},
\ee
and $w_j \ge c$ for each $j$. It can be shown, in fact, that $w_j\sim 1$.
Algorthms to compute the quadrature weights are discussed in \cite{quadconst}. \qed}
\end{rem}

\subsection{Smoothness classes and operators}\label{smoothsect}

First, we define a smoothness class on the sphere, comprising essentially of those functions which are a ``continuous ZF network'' of the form \eref{fundaidentity} below. In the sequel, if $a\in\CC$, we define
\be\label{invdef}
a^{[-1]}=\left\{\begin{array}{ll}
1/a, &\mbox{if $a\not=0$},\\
0,&\mbox{ if $a=0$}.
\end{array}\right.
\ee
We denote by $W_{q;\phi}^p$ the set of all $f\in X^p$ such that there exists $\mathcal{D}_\phi(f)\in X^p$ such that (cf. \cite{psdiff})
\be\label{dphidef}
\widehat{\mathcal{D}_\phi(f)}(\ell,k)=\hat{\phi}(\ell)^{[-1]}\hat{f}(\ell,k),
\qquad k=1,\cdots,d_\ell^q,\ \ell=0,1,\cdots.
\ee
Using \eref{sphereformalexpansion}, it is then clear that
\be\label{fundaidentity}
f(\x)=\int_{\SS^q}\phi(\x\cdot\y)\mathcal{D}_\phi(f)(\y)d\mu_q^*(\y), \qquad f\in W_{q;\phi}^p.
\ee

It is clear that the class $W_{q;\phi}^p$ contains all spherical polynomials. 
Lemma~\ref{sph_poly_deg_approx_lemma} can be used to show that the class of all $f\in L^p$ for which $\|f\|_p+ \|\mathcal{D}_\phi(f)\|_p \le 1$ is compact.

In the case when $\phi$ is positive definite; i.e.,  $\hat{\phi}(\ell)>0$ for all $\ell$, then the space $W_{q;\phi}^2$ is the reproducing kernel Hilbert space with $\phi$ as the reproducing kernel. 
Since the $L^2$ norm is blind to the signs of the expansion coefficients,  the theory of approximation in that norm is the same for ZF networks with the activation function $\phi$ and the function $\phi^+$ obtained from $\phi$ by replacing $\hat{\phi}(\ell)$'s by $|\hat{\phi}(\ell)|$'s; in particular,
$\|\mathcal{D}_\phi(f)\|_2=\|\mathcal{D}_{\phi^+}(f)\|_2$, and the theory in \cite{zfquadpap} can be used for studying approximation in $L^2$ norm.

 Our main interest in this paper  is approximation in the uniform norm as required in the applications to deep networks (cf. \cite{dingxuanpap}). 
Thus, the main technical challenge of this paper is that the function $\phi_\gamma(t)=|t|^{2\gamma+1}$ is not positive definite. In fact, it follows from Proposition~\ref{actcoeffprop} (used with $\alpha=q/2-1$) that $\widehat{\phi_\gamma}(2\ell+1)=0$ and with  $x!=\Gamma(x+1)$,
\be\label{gammafactor}
\widehat{\phi_\gamma}(2\ell)=
(-1)^\ell \frac{\cos(\pi\gamma)(q/2-1)!(2\gamma+1)!}{2^{2\gamma+1}\sqrt{\pi}} \frac{(\ell -\gamma-3/2)!}{(\ell+\gamma +q/2)!}. 
\ee 

\begin{figure}[h]
\begin{center}
\begin{minipage}{0.4\textwidth}
\begin{center}
\includegraphics[width=0.5\textwidth,keepaspectratio]{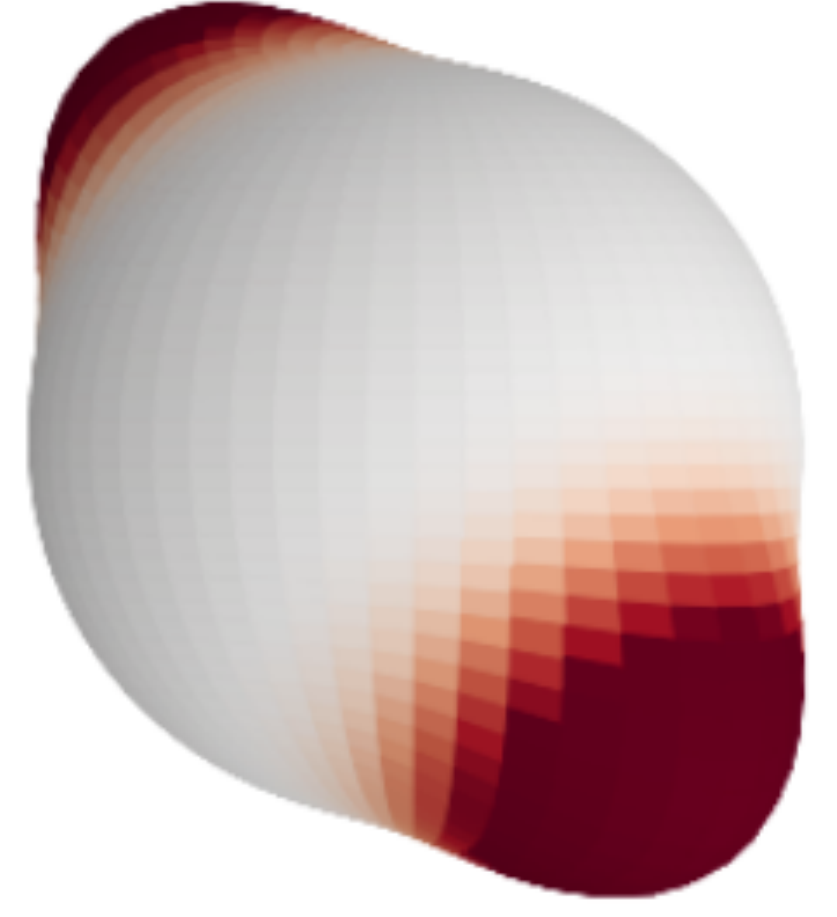} 
\end{center}
\end{minipage}
\begin{minipage}{0.4\textwidth}
\begin{center}
\includegraphics[width=0.5\textwidth,keepaspectratio]{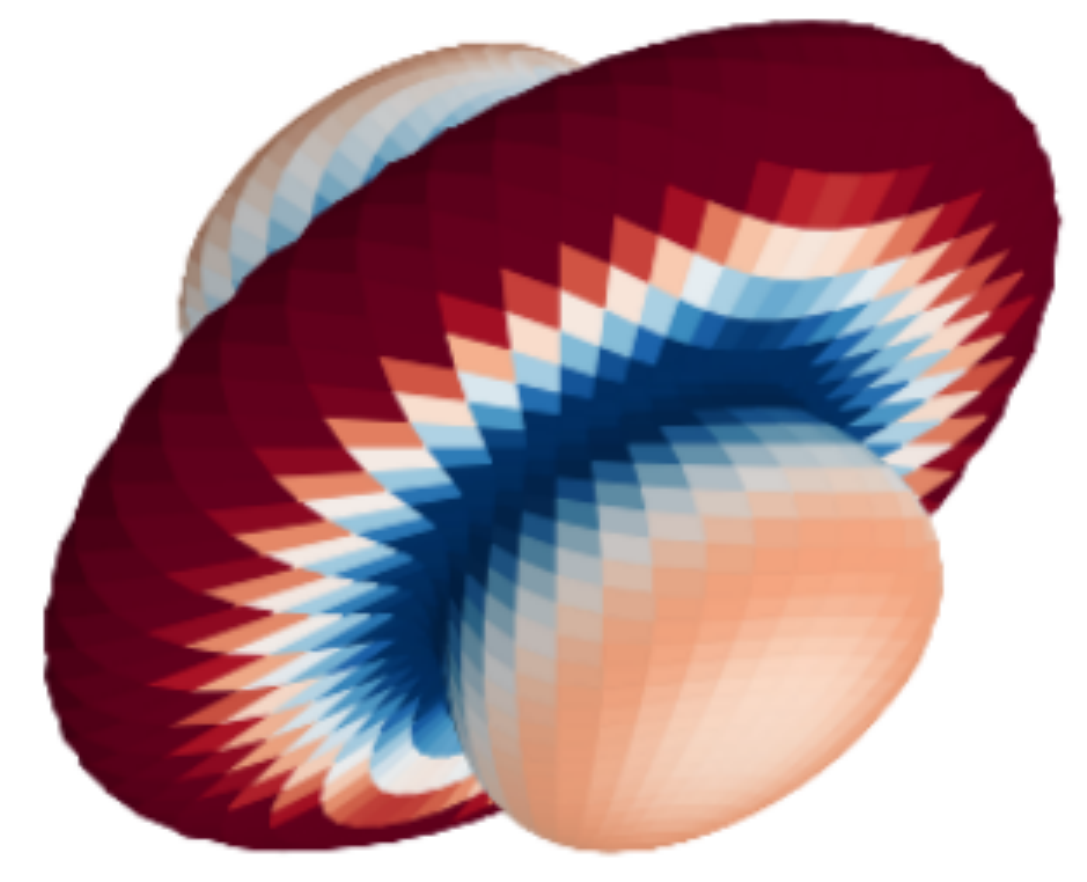} 
\end{center}
\end{minipage}
\end{center}
\caption{On the left, with $\x_0=(1,1,1)/\sqrt{3}$, the graph of $f(\x)=[(\x\cdot\x_0-0.1)_+]^8 + [(-\x\cdot\x_0-0.1)_+]^8$. On the right, the graph of $\mathcal{D}_{\phi_0}(f)$. Figure courtesy of D. Batenkov.}
\label{reluapproxfig}
\end{figure}

The fact that $\phi_\gamma$ is not a positive definite function on the sphere implies that $\mathcal{D}_{\phi_\gamma}$ is not a differential operator. 
As shown in Figure~\ref{reluapproxfig}, it is not even a local
operator; i.e., if a function $f$ is supported on a cap centered at a point $\x_0\in\SS^q$,   $\mathcal{D}_{\phi_\gamma}(f)$ is not  supported on this cap, but instead on  an equatorial region perpendicular to $\x_0$.

The lack of positive definiteness makes the theory in this paper quite different from that in \cite{zfquadpap} in norms other than the $L^2$ norm. The ultraspherical polynomials no longer play just the role of eigenfunctions of some operator,  enabling a generalization as in \cite{eignet}, but their relationship with other Jacobi polynomials plays a central role.
Moreover, the techniques from \cite{eignet} to obtain converse theorems no longer work. This leads us to conjecture that our results in Theorem~\ref{mainshallowtheo} and Corollary~\ref{phigammacor} are best possible in the sense of widths, but not in the sense of a converse theorem.

We now turn our attention to the conditions that we wish to require on $\{\hat{\phi}(\ell)\}_{\ell=0}^\infty$ instead of positivity. For a sequence $\bs{a}=\{a_\ell\}_{\ell=0}^\infty$, the forward differences are defined by
\be\label{fordiffdef}
\Delta a_\ell=\Delta^1 a_\ell=a_{\ell+1}-a_{\ell},\ \Delta^r a_\ell=\Delta(\Delta^{r-1} a_\ell), \ \ell\ge0, \ r\ge 2.
\ee

\begin{definition}\label{maskdef}
Let $s\in\RR$. A sequence  $\mathbf{b}=\{b_\ell\}_{\ell=0}^\infty$ of real numbers is called an $s$-sequence (written $\mathbf{b}\in \mathcal{B}(s)$) if 
\be\label{pseudodiffcond}
\sup_{\ell, r\ge 0} (\ell+1)^r|\Delta^r (\ell+1)^{s}b_\ell| \le c.
\ee
 The class $\mathcal{A}(s)$ consists of continuous, even functions $\phi :[-1,1]\to\RR$  such that $\{(-1)^\ell\hat{\phi}(2\ell)\}_{\ell=0}^\infty\in \mathcal{B}(s)$.
\end{definition}

In Corollary~\ref{phiexistcor}, we will show that for any sequence $\mathbf{b}\in \mathcal{B}(s)$, $s>(q+1)/2$, there exists a continuous, even function $\phi$ such that $\hat{\phi}(2\ell)=(-1)^\ell b_\ell$, $\ell=0,1,2,\cdots$, so that $\phi\in \mathcal{A}(s)$. We note that the condition \eref{pseudodiffcond} is satisfied if $\mathbf{b}$ satisfies an asymptotic expansion of the form
\be\label{tempasymp}
b_\ell=\ell^{-s}\sum_{j=0}^\infty \frac{d_j}{\ell^j}.
\ee
In particular, $\phi_\gamma\in \mathcal{A}((4\gamma+3+q)/2)$ if $\gamma>-1/2$ and $2\gamma+1$ is not an even integer (cf. \eref{actcoeff2}).

Finally, we define an operator and review its properties. Let $S>q+1$ be an integer, $h :[0,\infty)\to [0,1]$ be an $S$ times continuously differentiable function such that $h(t)=1$ if $0\le t\le 1/2$, $h(t)=0$ if $t\ge 1$.
We write
\be\label{usualkerndef}
\Phi_n(\x,\y)=\omega_{q-1}^{-1}\sum_{\ell=0}^\infty h(\ell/n) p_{2\ell}^{(q/2-1,q/2-1)}(1)p_{2\ell}^{(q/2-1,q/2-1)}(\x\cdot\y), \qquad n>0, \ \x\in\SS^q.
\ee
If $\nu$ is a measure on $\SS^q$, and $f\in L^1(\nu)$ is an even function, we define
\be\label{summopdef}
\sigma_n(\nu;f,\x)=\int_{\SS^q}f(\y)\Phi_n(\x,\y)d\nu(\y).
\ee 

If $1\le p\le\infty$,  $f\in L^p$ and $n>0$, we define
\be\label{polydegapproxdef}
E_{n,p}(f)=\min\{\|f-P\|_p : P\in \Pi_n^q\},
\ee
and write $X^p=\{f\in L^p : \disp\lim_{n\to\infty}E_{n,p}(f)=0\}$. Thus, $X^p=L^p$ if $1\le p<\infty$ and $C(\SS^q)$ if $p=\infty$. 

The following proposition summarizes important properties of the operator $\sigma_n$. The estimate \eref{sigmaopbd} is proved in \cite{modlpmz}, the estimate \eref{goodapprox} was proved in \cite{locsmooth} (with a different notation), the rest of the assertions can be verified easily from the definitions.

\begin{prop}\label{sigmaopprop}
Let $1\le p\le\infty$, $n\ge 1$, $\nu$ be an MZ quadrature measure of order $4n$, $f\in L^p$ be an even function. Then $\sigma_n(\nu;f)\in \Pi_{2n}^q$. If $P\in \Pi_n^q$ is an even polynomial then $\sigma_n(\nu;P)=P$. We have
\be\label{sigmaopbd}
\|\sigma_n(\nu;f)\|_p \le c\tn\nu\tn_{R,1/n}^{1/p'}\|f\|_{\nu;p},
\ee
and
\be\label{goodapprox}
E_{4n,p}(f)\le \|f-\sigma_n(\nu;f)\|_p \le c\tn\nu\tn_{R,1/n}^{1/p'}\min_{P\in\Pi_n^q}\|f-P\|_{\nu;p}.
\ee
Further,
\be\label{sigmacommute}
\sigma_n(\mu_q^*;\mathcal{D}_\phi(f))
=\mathcal{D}_\phi\left(\sigma_n(\mu_q^*;f)\right).
\ee
\end{prop}

\bhag{Main results}\label{mainsphapproxsect}
The results in this section are stated in a much greater generality than required for studying approximation on the Euclidean space, partly because the details are not any simpler in restricting ourselves to the case $\phi_\gamma(t)=|t|^{2\gamma+1}$ and the uniform norm. Accordingly, our main theorem, Theorem~\ref{mainshallowtheo} will be stated in a very general and abstract form.

In this section, let $\phi : [-1,1]\to\RR$ be an even, continuous function in $\mathcal{A}(s)$ for some $s>(q+1)/2$. Let   $\mu, \nu$ be  measures on $\SS^q$, $f\in L^p(\mu)$. We define for $n>0$, the (abstract) ZF network approximating $f$ (cf. \eref{zfnetexplicit}) by
\be\label{nnnetdef}
G_n(\phi,\mu, \nu;f,\x)
=\int_{\SS^q}\phi(\x\cdot\y)\mathcal{D}_\phi\left(\sigma_n(\mu;f)\right)(\y)d\nu(\y).
\ee
In this notation, $\phi$ is the activation function, $f$ is the target function, $\mu$ is the measure depending upon the available information about the target function $f$ (e.g., if the information is $\{f(\xi)\}_{\xi\in \tilde{C}}$, then $\mu$ is a discrete measure supported on $\tilde{C}$). The measure $\mu$ is required to be a quadrature measure of order $\sim n$ (which determines $n$) and $\nu$ is a quadrature measure of the same order supported on a judiciously chosen set of points. The size of the network is the same as the size of the support of $\nu$.

Before stating the main theorem, we first illustrate its implementation, and make some comments, explaining in particular why \eref{nnnetdef} defines an abstract ZF network.
 In Algorithm~1 below, we illustrate the construction of the network in the case when $p=\infty$, using a training data of the form $(\xi, f(\xi))$, $\xi\in\tilde{C}$, for a finite subset $\tilde{C}\subset \SS^q$.
\begin{algorithm}[h]\label{algfigure}
\caption{Construction of ZF network}
\begin{algorithmic}[1]
\item[{\rm a)}] \textbf{Input:} $(\xi, f(\xi))$, $\xi\in\tilde{C}$, for a finite subset $\tilde{C}\subset \SS^q$.
\item[{\rm b)}] Find an integer $N$, and weights $\tilde{w}_\xi$, $\xi\in\tilde{C}$, by solving the under-determined system of equations
$$
\sum_{\xi\in \tilde{C}} \tilde{w}_\xi Y_{\ell, j}(\xi)=\delta_{(\ell,j), (0,0)}, \qquad j=1,\cdots, d_\ell^q, \ \ell=0,\cdots, 4N,
$$
/* The measure $\mu$ in Theorem~\ref{mainshallowtheo} associates the mass $\tilde{w}_\xi$ with $\xi$, $\xi\in\tilde{C}$. It can be shown that $\tn\mu\tn_{R, 1/(4N)}$ is at most a constant times the largest  singular value of the matrix involved in the above system of equations. See Remark~\ref{quadconstrem} for further discussion.*/
\item[{\rm c)}] With $N$ as above, we use any known quadrature formula  i.e., a set of points $\C=\{\x_k\}$, and weights $w_k\ge 0$ such that
\be\label{quadratureexplicit}
\sum_{x_k\in\C} w_kP(\x_k)=\int_{\SS^q}Pd\mu_q^*, \qquad P\in \Pi^q_{4N}.
\ee
/* The measure $\nu$ in Theorem~\ref{mainshallowtheo} associates the mass $w_k$ with $\x_k$, $\x_k\in\C$. In view of Proposition~\ref{quadratureprop}(c), the measure $\nu$ is an MZ quadrature measure of order $4N$.*/
\item[{\rm d)}] Let
$$
\Psi_N(\x_k,\xi)=w_k\tilde{w}_\xi \sum_{\ell=0}^{N}\sum_{j=1}^{d_{2\ell}^q} h\left(\frac{\ell}{N}\right)\hat{\phi}(\ell)^{[-1]}Y_{2\ell,j}(\x_k)\overline{Y_{2\ell,j}(\xi)}, \qquad \x_k\in \C,\ \xi\in\tilde{C}.
$$
\item[{\rm e)}] \textbf{Output:}
\be\label{zfnetexplicit}
\mathbb{G}(\phi,\mu,\nu;f,\x)=\sum_{x_k\in\C}\left\{\sum_{\xi\in\tilde{C}}\Psi_N(\x_k,\xi)f(\xi)\right\}\phi(\x\cdot\x_k)= \sum_{\xi\in\tilde{C}}\left\{\sum_{x_k\in\C}\Psi_N(\x_k,\xi)\phi(\x\cdot\x_k)\right\}f(\xi).
\ee
\end{algorithmic}
\end{algorithm}

\begin{rem}\label{quadconstrem}
{\rm
The quantity $N$ in part (b) is a tunable parameter, akin to the regularization parameter in learning theory. A straightforward ``brute force'' way to make the choice is to start with the maximum possible value of $N$ given $|\tilde{C}|$, lower the value until the system has a solution, then carry out the rest of the algorithm for all values of $N$ 
from this value to $0$, and then choose the value that gives the best error.  Some heuristics for the choice of $N$  are:  to minimize $\sum_{\xi\in \tilde{C}} |\tilde{w}_\xi|^2$ or to control the residual error or  the condition number. 

The construction of a quadrature formula as in part (b) above is given in \cite{quadconst}. In this paper, the use of this formula in approximation of functions on the sphere is illustrated using several benchmark examples. An application in the case of a $4$-dimensional sphere for the classification of drussen in age related amacular disease is discussed in \cite{compbio}.
\qed}
\end{rem}
\begin{rem}\label{networkrem}
{\rm
The first formula in \eref{zfnetexplicit} shows that $\mathbb{G}$ is a ZF network, whose  coefficients can be computed as a linear combination of the training data. Also, the centers of this network are \textbf{independent of the training data.}
The second formula in \eref{zfnetexplicit} shows that the network can be constructed by treating the training data as coefficients of a set of fixed networks constructed independently of the target function; based entirely on the locations of the sampling points.
\qed}
\end{rem}

\begin{rem}\label{sphrotinvariancerem}
{\rm
In view of \eref{addformula}, if $U$ is a rotation of $\RR^{q+1}$, and $f_U(\x)=f(U\x)$ for $\x\in\SS^q$, then the rotation invariance of $\mu_q^*$ implies that for integer $\ell=0,1,\cdots$, one has
$$
\sum_{j=1}^{d_\ell^q}\hat{f_U}(\ell,j)Y_{\ell, j}(\x)=\omega_{q-1}^{-1}p_\ell^{(q/2-1,q/2-1)}(1)\int_{\SS^q}f(U\y)p_\ell^{(q/2-1,q/2-1)}(\x\cdot\y )d\mu_q^*(\y)=\sum_{j=1}^{d_\ell^q}\hat{f}(\ell,j)Y_{\ell, j}(U\x).
$$
The rotation invariance implies also that if $\{\x_k\}$ satisfies \eref{quadratureexplicit}, then so does the system $\{U\x_k\}$, with the same weights. Therefore, using the definition \eref{zfnetexplicit} once with $f_U$ in place of $f$, and once with $\{U\x_k\}$ in place of $\{\x_k\}$,
it is easy to deduce that
\be\label{sphrotinvariance}
G_{2^n}(\phi,\mu_q^*;\nu;f_U,\x)=G_{2^n}(\phi,\mu_q^*;\nu;f,U\x), \qquad \x\in\SS^q.
\ee
\qed}
\end{rem}

Our main theorem is the following (cf. \cite[Theorem~3.1]{zfquadpap}).

\begin{theorem}\label{mainshallowtheo}
Let $s>(q+1)/2$, $\phi\in\mathcal{A}(s)$, $1\le p\le \infty$, $f\in W_{q;\phi}^p$, $n\ge 1$ and $\nu$ be an MZ quadrature measure of order $2^{n+2}$.  
Then
\be\label{shallowdegapprox}
\|f-G_{2^n}(\phi, \mu_q^*,\nu;f)\|_p \le c2^{-n(s-(q-1)/2)}(1+\tn\nu\tn_{R,1/2^n})\|\mathcal{D}_\phi(f)\|_p.
\ee
Moreover, the coefficients in the network  satisfy
\be\label{sphcoefftheory}
\int_{\SS^q}|\mathcal{D}_\phi\left(\sigma_n(\mu_q^*;f)\right)(\y)|d|\nu|(\y)\le c\|
\mathcal{D}_\phi(f)\|_1.
\ee
If $p=\infty$, and $\mu$ is an MZ quadrature measure of order $2^{n+2}$ then
\be\label{sampleshallowdegapprox}
\|f-G_{2^n}(\phi, \mu,\nu;f)\|_\infty \le c2^{-n(s-(q-1)/2)}\tn\mu\tn_{R,1/2^n}(1+\tn\nu\tn_{R,1/2^n})\|\mathcal{D}_\phi(f)\|_\infty.
\ee
\end{theorem}

We state a corollary for the special case $\phi_\gamma(t)=|t|^{2\gamma+1}$, obtained by noting from \eref{actcoeff2} that $\phi_\gamma\in \mathcal{A}((4\gamma+3+q)/2)$.
\begin{cor}\label{phigammacor}
Let $\gamma>-1/2$, $2\gamma+1$ not be an even integer. Let $1\le p\le \infty$, $f\in W_{q;\phi_\gamma}^p$, $n\ge 1$ and $\nu$ be an MZ quadrature measure of order $2^{n+2}$.  
Then
\be\label{phigammadegapprox}
\|f-G_{2^n}(\phi_\gamma, \mu_q^*,\nu;f)\|_p \le c2^{-2n(\gamma+1)}(1+\tn\nu\tn_{R,1/2^n})\|\mathcal{D}_{\phi_\gamma}(f)\|_p.
\ee
Moreover, the coefficients in the network  satisfy
\be\label{phigammasphcoefftheory}
\int_{\SS^q}|\mathcal{D}_\phi\left(\sigma_n(\mu_q^*;f)\right)(\y)|d|\nu|(\y)\le c\|
\mathcal{D}_{\phi_\gamma}(f)\|_1.
\ee
If $p=\infty$, and $\mu$ is an MZ quadrature measure of order $2^{n+2}$ then
\be\label{phigammasampledegapprox}
\|f-G_{2^n}(\phi_\gamma, \mu,\nu;f)\|_\infty \le c2^{-2n(\gamma+1)}\tn\mu\tn_{R,1/2^n}(1+\tn\nu\tn_{R,1/2^n})\|\mathcal{D}_{\phi_\gamma}(f)\|_\infty.
\ee
\end{cor}

\begin{rem}\label{dyadicrem}
{\rm
Although the theorem and its corollary are stated with dyadic integers $2^n$, this is only for the convenience of the proof. Since for any integer $m$, we can always find $n$ with $2^{n-1}\le m\le 2^n$, in practice, we may use other integers as well without affecting the degree of approximation except for the constants involved.
\qed}
\end{rem}

\begin{rem}\label{relurem}
{\rm
In the case of the ReLU activation function $t\mapsto |t|$,  the right hand sides of the estimates \eref{phigammadegapprox} and \eref{phigammasampledegapprox} are both $\O(2^{-2n})$, a substantial improvement on the result announced in \cite{dingxuanpap}. In terms of the number $M$ of parameters involved in these ZF networks,  the estimates are $\O(M^{-2/q})$. 
\qed}
\end{rem}

\begin{rem}\label{coeffrem}
{\rm
The left hand side of \eref{sphcoefftheory} (respectively, \eref{phigammasphcoefftheory}) expresses a weighted sum of the coefficients of the networks involved. Since the right hand side is independent of $n$, these estimates show that  this sum can be bounded independently of the number of (Fourier) data available and size of the networks.
\qed}
\end{rem}
\bhag{Technical preparation}\label{prepsecct}
In this section, we review some facts about Jacobi polynomials, and certain kernels defined with these. In the sequel, we denote $\Gamma(z+1)$ by $z!$. 

For $\alpha, \beta>-1$, $x\in (-1,1)$ and integer $\ell\ge 0$, the Jacobi polynomials $p_\ell^{(\alpha,\beta)}$ are defined by the Rodrigues' formula \cite[Formulas~(4.3.1), (4.3.4)]{szego}
\be\label{rodrigues}
(1-x)^\alpha(1+x)^\beta p_\ell^{(\alpha,\beta)}(x)=\left\{\frac{2\ell+\alpha+\beta+1}{2^{\alpha+\beta+1}}\frac{\ell!(\ell+\alpha+\beta)!}{(\ell+\alpha)!(\ell+\beta)!}\right\}^{1/2}\frac{(-1)^\ell}{2^\ell \ell!}\frac{d^\ell}{dx^\ell}\left((1-x)^{\ell+\alpha}(1+x)^{\ell+\beta}\right).
\ee
Each $p_\ell^{(\alpha,\beta)}$ is a polynomial of degree $\ell$ with positive leading coefficient,  one has the orthogonality relation
\be\label{jacobiortho}
\int_{-1}^1 p_\ell^{(\alpha,\beta)}(x)p_j^{(\alpha,\beta)}(x)(1-x)^\alpha(1+x)^\beta=\delta_{\ell,j},
\ee
and
\be\label{pkat1}
p_\ell^{(\alpha,\beta)}(1)=\left\{\frac{2\ell+\alpha+\beta+1}{2^{\alpha+\beta+1}}\frac{\ell!(\ell+\alpha+\beta)!}{(\ell+\alpha)!(\ell+\beta)!}\right\}^{1/2}\frac{(\ell+\alpha)!}{\alpha!\ell!} \sim \ell^{\alpha+1/2}.
\ee
It follows that $p_\ell^{(\alpha,\beta)}(-x)=(-1)^\ell p_\ell^{(\beta,\alpha)}(x)$. In particular,  $p_{2\ell}^{(\alpha,\alpha)}$ is an even polynomial, and $p_{2\ell+1}^{(\alpha,\alpha)}$ is an odd polynomial.  We note (cf. \cite[Theorem~4.1]{szego}) that
\bea
p_{2\ell}^{(\alpha,\alpha)}(x)&=&2^{\alpha/2+1/4}p_\ell^{(\alpha,-1/2)}(2x^2-1)=2^{\alpha/2+1/4}(-1)^\ell p_\ell^{(-1/2,\alpha)}(1-2x^2),\label{evenjacobi}\\
p_{2\ell+1}^{(\alpha,\alpha)}(x)&=&2^{\alpha/2+1/2}xp_\ell^{(\alpha,1/2)}(2x^2-1)=2^{\alpha/2+1/2}(-1)^\ell xp_\ell^{(1/2,\alpha)}(1-2x^2).
\label{oddjacobi}
\eea

Next, we discuss the localization properties of some kernels associated with Jacobi polynomials. 
 We record the following theorem, which follows easily from  \cite[Lemma~4.10]{locjacobi}.
 
\begin{theorem}\label{locjacobitheo}
Let $\alpha,\beta\ge -1/2$, $S>\max(\alpha+3/2, \alpha-\beta+1)$, $n\ge 1$ be  integers, $\bs{a}=\{a_\ell\}_{\ell=0}^\infty$ be a sequence, such that $a_\ell=0$ for all $\ell\ge n$.
Then for $\theta\in [0,\pi]$,
\bea\label{localjacobiest}
\lefteqn{\left|\sum_{\ell=0}^\infty a_\ell p_\ell^{(\alpha,\beta)}(1)p_\ell^{(\alpha,\beta)}(\cos\theta)\right|}\nonumber\\
&\le& c_1\disp\sum_{\ell=0}^\infty (\ell+1)^{2\alpha+2}\sum_{r=1}^K \frac{1}{\max(1, (\ell+1)\theta)^S)}(\ell+1)^{r-1}|\Delta^r a_\ell|,
\eea
where $K\ge \max(S-\alpha-1/2, S-\alpha+\beta)$ is an integer ($\ge 1$).
\end{theorem}

\begin{Proof}\ 
In this proof only, let
$$
\Psi(\theta)=\sum_{\ell=0}^\infty a_\ell p_\ell^{(\alpha,\beta)}(1)p_\ell^{(\alpha,\beta)}(\cos\theta).
$$
In \cite[Lemma~4.10]{locjacobi} (used with $y=\cos\theta$), we have proved that for integer $K\ge 1$, 
\be\label{endptests1}
|\Psi(\theta)| \le c\left\{\begin{array}{ll}
\sum_{\ell=0}^\infty \min\left((\ell+1)^2, (1-\cos\theta)^{-1}\right)^{\alpha/2+K/2+1/4}\times & \\
\disp\qquad\times\sum_{m=0}^{K-1}(\ell+1)^{\alpha+1/2-m}|\Delta^{K-m}a_\ell|, 
& \mbox{if $0\le \theta\le \pi/2$,}\\
\disp\sum_{\ell=0}^\infty (\ell+1)^{\alpha+\beta+1} \sum_{m=0}^{K-1}(\ell+1)^{-m}|\Delta^{K-m}a_\ell|, 
& \mbox{if $\pi/2<\theta\le \pi$.}
\end{array}\right.
\ee
First, let $0\le \theta\le \pi/2$. A change in the summation index $m$ to $K-r$ yields
$$
\sum_{m=0}^{K-1}(\ell+1)^{\alpha+1/2-m}|\Delta^{K-m}a_\ell|
=(\ell+1)^{\alpha+3/2-K}\sum_{r=1}^K (\ell+1)^{r-1}|\Delta^r a_\ell|.
$$
Since $1-\cos\theta\sim \theta^2$, and $K+\alpha+1/2\ge S$, the first estimate in 
\eref{endptests1} leads to \eref{localjacobiest}. If $\pi/2<\theta\le \pi$, then  $1-\cos\theta\ge 1$. Therefore, a similar change of the index of summation in the second estimate in \eref{endptests1} leads to   \eref{localjacobiest}.
\end{Proof}

We will use heavily the following consequence of Theorem~\ref{locjacobitheo}.

\begin{theorem}\label{locjacobiusetheo}
Let $\alpha\ge -1/2$,  $S>\alpha+3/2$ be an integer, $G :[0,\infty)\to [0,1]$ be $S$ times continuously differentiable,  $G(t)=0$ for $t$ in a neighborhood of $0$, $G(t)=0$ if $t\ge 1$. Let $s\in\RR$, $\{b_\ell\}_{\ell=0}^\infty\in \mathcal{B}(s)$.
Then
\bea
\left|\sum_{\ell=0}^\infty b_\ell G\left(\frac{\ell}{n}\right)p_{2\ell}^{(\alpha,\alpha)}(1)p_{2\ell}^{(\alpha,\alpha)}(\cos\theta)\right|
&\le& c_1(G)\frac{n^{-s+2\alpha+2}}{\max(1, (n\theta)^{S}, (n(\pi-\theta))^{S})}\label{locjacobiuseful}\\
\left|\sum_{\ell=0}^\infty (-1)^\ell b_\ell G\left(\frac{\ell}{n}\right)p_{2\ell}^{(\alpha,\alpha)}(1)p_{2\ell}^{(\alpha,\alpha)}(\cos\theta)\right|
&\le& c_2(G)\frac{n^{-s+\alpha+3/2}}{\max(1, (n|\pi/2-\theta|)^{S})}\label{locjacobi_veryuseful}
\eea
\end{theorem}

\vskip 2ex
\begin{rem}\label{nonpdrem}
{\rm
We note that taking $\alpha=q/2-1$ and interpreting $\cos\theta=\x\cdot\y$, the non-positive definite kernel in \eref{locjacobi_veryuseful} is localized as a function of $\y$ around an equator perpendicular to $\x$. This explains also the behavior exhibited in Figure~\ref{reluapproxfig}. 
\qed}
\end{rem}
\vskip 2ex\noindent
\textsc{Proof of Theorem~\ref{locjacobiusetheo}.}\\
We note first that the assumption \eref{pseudodiffcond}  implies 
 (cf. \cite[Lemma~4.3(a), (b)]{quadconst}) that for any $t\in\RR$,
$$
\sum_{\ell=0}^\infty \sum_{r=1}^S (\ell+1)^{t+r-1}|\Delta^r (b_\ell G(\ell/n))| \le c\sum_{\ell=0}^\infty \sum_{r=1}^S (\ell+1)^{t-s+r-1}|\Delta^r (G(\ell/n))|.
$$
Using the mean value theorem, and the fact that $G$ is supported on $[a,1]$ for some $a>0$, we conclude (cf. \cite[Proof of Theorem~3.1]{locjacobi}) that for any $t\in\RR$,
\be\label{pf1eqn1}
\sum_{\ell=0}^\infty \sum_{r=1}^S (\ell+1)^{t+r-1}|\Delta^r (b_\ell G(\ell/n))| \le cn^{t-s}.
\ee
 Therefore, \eref{localjacobiest} in this case (keeping in mind that $2\theta\in [0,2\pi]$ while \eref{localjacobiest} applies to $\theta\in [0,\pi]$) leads to
 \be\label{pf1eqn2}
\left|\sum_{\ell=0}^\infty b_\ell G\left(\frac{\ell}{n}\right)p_\ell^{(\alpha,\beta)}(1)p_\ell^{(\alpha,\beta)}(\cos(2\theta))\right| \le   c_1(G)\frac{n^{-s+2\alpha+2}}{\max(1, (n\theta)^{S}, (n(\pi-\theta))^{S})}.
\ee
We use this estimate with $\beta=-1/2$, and recall the first equation in \eref{evenjacobi} to deduce \eref{locjacobiuseful}. 

To obtain \eref{locjacobi_veryuseful}, we deduce first using the Leibnitz formula for differences  that   
$$\{b_\ell p_{2\ell}^{(\alpha,\alpha)}(1)/p_\ell^{(-1/2,\alpha)}(1)\}_{\ell=0}^\infty\in \mathcal{B}(s-\alpha-1/2).
$$ 
Therefore,  we use the second equality in \eref{evenjacobi}, and \eref{pf1eqn2} with $-1/2$ in place of $\alpha$, $\alpha$ in place of $\beta$, $s-\alpha-1/2$ in place of $s$, to conclude that
\begin{eqnarray*}
\lefteqn{\left|\sum_{\ell=0}^\infty (-1)^\ell b_\ell G\left(\frac{\ell}{n}\right)p_{2\ell}^{(\alpha,\alpha)}(1)p_{2\ell}^{(\alpha,\alpha)}(\cos\theta)\right|}\\
&=& 2^{\alpha/2+1/4}\left|\sum_{\ell=0}^\infty \frac{b_\ell p_{2\ell}^{(\alpha,\alpha)}(1)}{p_\ell^{(-1/2,\alpha)}(1)}G\left(\frac{\ell}{n}\right)p_\ell^{(-1/2,\alpha)}(1)p_\ell^{(-1/2,\alpha)}(\cos(\pi-2\theta))\right|\\
& \le& c_2(G)\frac{n^{-s+\alpha+3/2}}{\max(1, (n|\pi/2-\theta|)^{S})}.
\end{eqnarray*}
\qed

In the remainder of this paper, we overload the notation and write
\be\label{unikerndef}
\tilde{\Phi}_n(H, \mathbf{b}; t)= \omega_{q-1}^{-1}\sum_{\ell=0}^\infty (-1)^\ell b_\ell H\left(\frac{\ell}{n}\right)p_{2\ell}^{(\alpha,\alpha)}(1)p_{2\ell}^{(\alpha,\alpha)}(t), \qquad t\in [-1,1].
\ee
 We set $g(t)=h(t)-h(2t)$, and note that for all $n\ge 1$, $t\in [0,\infty)$,
\be\label{hgrel}
h(t)+\sum_{m=1}^n g\left(\frac{t}{2^m}\right)=h\left(\frac{t}{2^n}\right), \qquad h\left(\frac{t}{2^n}\right)+\sum_{m=n+1}^\infty g\left(\frac{t}{2^m}\right)=1.
\ee
\begin{cor}\label{phiexistcor}
Let   $s>\alpha+3/2$, and $\mathbf{b}\in \mathcal{B}(s)$. Then there exists a continuous, even function $\phi : [-1,1]\to\RR$ such that
\be\label{phiexist}
\int_{-1}^1 p_{2\ell}^{(\alpha,\alpha)}(t)\phi(t)(1-t^2)^\alpha dt =(-1)^\ell b_\ell 
p_{2\ell}^{(\alpha,\alpha)}(1),
\ee
and
\be\label{phidegapprox}
\left\|\phi- \tilde{\Phi}_{2^n}(h, \mathbf{b};\circ)\right\|_{\infty, [-1,1]} \le c2^{-n(s-\alpha-3/2)}.
\ee
\end{cor}

\begin{Proof}\ 

In view of \eref{locjacobi_veryuseful}, we see that for $n\ge 0$,
$$
\sum_{j=n+1}^\infty \left\|\tilde{\Phi}_{2^j}(g, \mathbf{b}; \circ)\right\|_{\infty;[-1,1]} \le c\sum_{j=n+1}^\infty 2^{-j(s-\alpha-3/2)}<c2^{-n(s-\alpha-3/2)}.
$$
Using \eref{hgrel} (with $j$ in place of $t$), this leads easily to \eref{phidegapprox}, and hence, to the rest of the assertions of this corollary.
\end{Proof}

\bhag{Proof of of Theorem~\ref{mainshallowtheo}}\label{sphapproxsect}

The fundamental approach in the proof of Theorem~\ref{mainshallowtheo} mimics that of \cite[Theorem~3.1, equivalently, Theorem~6.1]{eignet}. 
Thus, we first prove the theorem when the target function is a spherical polynomial (Lemma~\ref{polyapproxlemma}). 
 The other main step is to obtain a degree of approximation of the actual target function by spherical polynomials in terms of the operator $\mathcal{D}_\phi$ (Lemma~\ref{sph_poly_deg_approx_lemma}). 
 The proof of Theorem~\ref{mainshallowtheo} is then a simple calculation combining the results of these two main steps. 
 A crucial step in these proofs is Proposition~\ref{fundaprop}.
  However, the details are very different because of the fact that the activation function is not positive definite. It will be surprising if these details can be worked out in the case of a general manifold in the same way.

We now start this program with two lemmas preparatory for Proposition~\ref{fundaprop}. The first of these encapsulates the differences needed from the proof of the corresponding \cite[Corollary~5.1]{eignet} because of the fact that the kernels are not localized around their centers but around equators around these centers.

\begin{lemma}\label{altkernlemma}
Let $0<d\le 1$, $\nu\in\mathcal{R}_d$, $s\in\RR$, $\mathbf{b}\in \mathcal{B}(s)$. Then for $n\ge 1$,
\be\label{altkernest}
\int_{\SS^q}|\tilde{\Phi}_n(g, \mathbf{b};\x\cdot\y)|d|\nu|(\y)\le cn^{-s+(q-1)/2}(1+nd)\tn\nu\tn_{R,d}, \qquad \x\in\SS^q.
\ee
\end{lemma}

\begin{Proof}\ 
Since the kernel $\tilde{\Phi}_n$ is an even function of $\x$ and $\y$, there is no loss of generality in assuming that $\nu$ is an even measure, and $\tn\nu\tn_{R,d}=1$.
In this proof only, let $\x\in\SS^q_+$ be fixed, and for $r,\rho\ge 0$,
\be\label{pf2eqn1}
A_{r,\rho}=\{\y\in\SS^q : \x\cdot\y=\cos\theta, \ r<\pi/2-\theta\le \rho,\ \theta\ge \pi/4\}.
\ee
We estimate first $|\nu|(A_{r,\rho})$. If $\rho-r > d$, then $A_{r,\rho}$ is covered by $c(\rho-r)^{-(q-1)}$ caps of radius $\sim \rho-r$ (i.e., volume $\sim (\rho-r)^q$) each. Therefore, 
$|\nu|(A_{r,\rho})\le c(\rho-r)$. If $\rho-r \le d$, then $A_{r,\rho}$ is covered by $d^{-(q-1)}$ caps of radius $\sim d$ each. Therefore, 
$|\nu|(A_{r,\rho})\le cd$. Thus,
\be\label{pf2eqn2}
|\nu|(A_{r,\rho})\le c(\rho-r)\left(1+\frac{d}{\rho-r}\right).
\ee

Next, we recall \eref{locjacobi_veryuseful} with $(q-2)/2$ in place of $\alpha$, and $\x\cdot\y=\cos\theta$:
\be\label{pf2eqn3}
|\tilde{\Phi}_n(g, \mathbf{b};\x\cdot\y)|\le c\frac{n^{-s+(q+1)/2}}{\max(1, |n(\pi/2-\theta)|^S)}.
\ee
It follows immediately from \eref{pf2eqn3} and \eref{pf2eqn2} that
\be\label{pf2eqn4}
\int_{A_{0,\pi/n}}|\tilde{\Phi}_n(g, \mathbf{b};\x\cdot\y)|d|\nu|(\y) \le cn^{-s+(q+1)/2}|\nu|(A_{0,\pi/n})\le cn^{-s+(q-1)/2}(1+nd),
\ee
and
\be\label{pf2eqn5}
\int_{\y\in\SS^q: \theta\le \pi/4} |\tilde{\Phi}_n(g, \mathbf{b};\x\cdot\y)|d|\nu|(\y) \le cn^{-s+(q+1)/2-S}|\nu|(\SS^q)\le cn^{-s+(q+1)/2-S}.
\ee
Finally, using \eref{pf2eqn3} and \eref{pf2eqn2} again,
\begin{eqnarray*}
\int_{A_{\pi/n,\pi/2}} |\tilde{\Phi}_n(g, \mathbf{b};\x\cdot\y)|d|\nu|(\y) &\le& cn^{-s+(q+1)/2-S}\sum_{k=0}^\infty \int_{A_{2^k\pi/n,2^{k+1}\pi/n}}\frac{d|\nu|(\y)}{|\pi/2-\theta|^S}\\
&\le& cn^{-s+(q+1)/2-S}\sum_{k=0}^\infty (2^k\pi/n)^{-S}|\nu|(A_{2^k\pi/n,2^{k+1}\pi/n})\\
& \le& cn^{-s+(q-1)/2}\sum_{k=0}^\infty 2^{-k(S-1)}(1+nd/2^k) \le cn^{-s+(q-1)/2}(1+nd).
\end{eqnarray*}
Together with \eref{pf2eqn4} and \eref{pf2eqn5}, this leads to \eref{altkernest}.
\end{Proof}

The following lemma follows easily from Lemma~\ref{altkernlemma}.

\begin{lemma}\label{altoplemma}
Let $0<d\le 1$, $\nu\in\mathcal{R}_d$, $\mathbf{b}\in \mathcal{B}(s)$, $1\le p\le\infty$, $F\in L^p(\nu)$. Then for $n\ge 1$,
\be\label{altopest}
\left\| \int_{\SS^q}\tilde{\Phi}_n(g, \mathbf{b};\circ\cdot\y)F(\y)d\nu(\y)\right\|_p \le
cn^{-s+(q-1)/2}\left((1+nd)\tn\nu\tn_d\right)^{1/p'}\|F\|_{\nu;p}.
\ee
\end{lemma}
\begin{Proof}
The estimate \eref{altopest} is clear from \eref{altkernest} for $F\in L^\infty(\nu)$. Applying \eref{altkernest} with $\mu_q^*$ in place of $\nu$, (so that we may choose $d=0$), we get
\be\label{pf3eqn2}
\int_{\SS^q}|\tilde{\Phi}_n(g, \mathbf{b};\x\cdot\y)|d\mu_q^*(\y)\le cn^{-s+(q-1)/2}.
\ee
Using Fubini's theorem, it is then easy to see that \eref{altopest} holds for $F\in L^1(\nu)$. The general case follows from the Riesz-Thorin interpolation theorem \cite[Theorem~1.1.1]{bergh}.
\end{Proof}

The following proposition is the analogue of \cite[Proposition~5.2]{eignet}.

\begin{prop}\label{fundaprop}
Let $s>(q+1)/2$, $\phi\in \mathcal{A}(s)$, $b_\ell=(-1)^\ell\hat{\phi}(2\ell)$,  $n\ge 1$ be an integer, $0<d\le 1$, $\nu\in \mathcal{R}_d$, $1\le p\le\infty$, $F\in L^p(\nu)$. For $m\ge \log_2(1/d)$, let
\be\label{umdef}
U_m(\nu;F,\x)=\int_{\SS^q}\left\{\phi(\x\cdot\y)-\tilde{\Phi}_{2^m}(h, \mathbf{b}; \x\cdot\y)\right\}F(\y)d\nu(\y), \qquad \x\in\SS^q.
\ee
Then
\be\label{umest}
\|U_m(\nu;F)\|_p \le c2^{-m(s+1/p-(q+1)/2)}(d\tn\nu\tn_{R,d})^{1/p'}\|F\|_{\nu;p}.
\ee
\end{prop}

\begin{Proof}\ 
Since $m\ge \log_2(1/d)$, we see that $2^jd \ge 1$ for all integers $j\ge m$. Consequently, \eref{altopest} used with  $2^j$ in place of $n$ shows that for $j\ge m$,
\begin{eqnarray*}
\left\| \int_{\SS^q}\tilde{\Phi}_{2^j}(g, \mathbf{b};\circ\cdot\y)F(\y)d\nu(\y)\right\|_p &\le&
c2^{-j(s-(q-1)/2)}\left((1+2^jd)\tn\nu\tn_d\right)^{1/p'}\|F\|_{\nu;p}\\
&\le& c2^{-j(s-(q+1)/2+1/p)}(d\tn\nu\tn_d)^{1/p'}\|F\|_{\nu;p}.
\end{eqnarray*}
Therefore (cf. \eref{hgrel}),
\be\label{pf3eqn4}
\|U_m(\nu;F)\|_p\le \sum_{j=m+1}^\infty \left\| \int_{\SS^q}\tilde{\Phi}_{2^j}(g, \mathbf{b};\circ\cdot\y)F(\y)d\nu(\y)\right\|_p \le c2^{-m(s+1/p-(q+1)/2)}(d\tn\nu\tn_{R,d})^{1/p'}\|F\|_{\nu;p}.
\ee
\end{Proof}

\begin{lemma}\label{polyapproxlemma}
Let $n\ge 0$, $P\in \Pi_{2^{n+1}}^q$ be an even polynomial, $\nu$ be an MZ quadrature measure of order $2^{n+2}$, $s>(q+1)/2$, $\phi\in\mathcal{A}(s)$. Then for $1\le p\le\infty$,
\be\label{polyapproxest}
\left\|P-\int_{\SS^q}\phi(\circ\cdot\y)\mathcal{D}_\phi(P)(\y)d\nu(\y)\right\|_p \le c2^{-n(s-(q-1)/2)}(1+\tn\nu\tn_{R,1/2^n})\|\mathcal{D}_\phi(P)\|_p.
\ee
\end{lemma}
\begin{Proof}\ 
In this proof, let $b_\ell=(-1)^\ell \hat{\phi}(2\ell)$, so that $\mathbf{b}\in \mathcal{B}(s)$. Let $\x\in \SS^q$. Since $\nu$ is an MZ quadrature measure of order $2^{n+2}$, and $\mathcal{D}_\phi(P), \tilde{\Phi}_{2^n}(h, \mathbf{b}; \x\cdot\circ)\in \Pi_{2^{n+1}}^q$, we have
$$
\int_{\SS^q}\tilde{\Phi}_{2^n}(h, \mathbf{b}; \x\cdot\y)\mathcal{D}_\phi(P)(\y)d\nu(\y)=\int_{\SS^q}\tilde{\Phi}_{2^n}(h, \mathbf{b}; \x\cdot\y)\mathcal{D}_\phi(P)(\y)d\mu_q^*(\y).
$$
Hence, \eref{fundaidentity} implies that
\begin{eqnarray*}
\lefteqn{P(\x)-\int_{\SS^q}\phi(\x\cdot\y)\mathcal{D}_\phi(P)(\y)d\nu(\y)}\\
&=&\int_{\SS^q} \phi(\x\cdot\y)\mathcal{D}_\phi(P)(\y)d\mu_q^*(\y)
-\int_{\SS^q}\phi(\x\cdot\y)\mathcal{D}_\phi(P)(\y)d\nu(\y)\\
&=& \int_{\SS^q} \left\{\phi(\x\cdot\y)-\tilde{\Phi}_{2^n}(h, \mathbf{b}; \x\cdot\y)\right\}\mathcal{D}_\phi(P)(\y)d\mu_q^*(\y)\\
&&\qquad
-\int_{\SS^q} \left\{\phi(\x\cdot\y)-\tilde{\Phi}_{2^n}(h, \mathbf{b}; \x\cdot\y)\right\}\mathcal{D}_\phi(P)(\y)d\nu(\y)\\
&=&U_n(\mu_q^*;\mathcal{D}_\phi(P),\x)-U_n(\nu;\mathcal{D}_\phi(P),\x).
\end{eqnarray*}
Thus,
$$
\left\|P-\int_{\SS^q}\phi(\circ\cdot\y)\mathcal{D}_\phi(P)(\y)d\nu(\y)\right\|_p \le 
\|U_n(\mu_q^*;\mathcal{D}_\phi(P))\|_p + \|U_n(\nu;\mathcal{D}_\phi(P))\|_p.
$$
We now use Proposition~\ref{fundaprop} twice, once with $\mu_q^*$ in place of $\nu$, and once with $\nu$, and (in both cases), with $\mathcal{D}_\phi(P)$ in place of $F$, $2^{-n}$ in place of $d$ to deduce that
$$
\left\|P-\int_{\SS^q}\phi(\circ\cdot\y)\mathcal{D}_\phi(P)(\y)d\nu(\y)\right\|_p \le c2^{-n(s-(q-1)/2)}\left\{\|\mathcal{D}_\phi(P)\|_p+\tn\nu\tn_{R,1/2^n}^{1/p'}\|\mathcal{D}_\phi(P)\|_{\nu;p}\right\}.
$$
Since $\nu$ is a regular measure of order $2^n$, Proposition~\ref{mzequivprop}(c) now yields
\eref{polyapproxest}.
\end{Proof}

\begin{lemma}\label{sph_poly_deg_approx_lemma}
Let $s>(q+1)/2$, $\phi\in\mathcal{A}(s)$, $1\le p\le \infty$, $f\in W_{q;\phi}^p$, $n\ge 1$
Then
\be\label{mustardegapprox}
\|f-\sigma_{2^n}(\mu_q^*;f)\|_p \le c2^{-n(s-(q-1)/2)}\|\mathcal{D}_\phi(f)\|_p.
\ee
In the case when $p=\infty$, and $\mu$ is an MZ quadrature measure of order $2^{n+2}$, we have
\be\label{mudegapprox}
\|f-\sigma_{2^n}(\mu;f)\|_\infty \le c2^{-n(s-(q-1)/2)}\tn\mu\tn_{R, 2^{-n}}\|\mathcal{D}_\phi(f)\|_\infty.
\ee
\end{lemma}

\begin{Proof}\ 
In this proof, let $b_\ell=(-1)^\ell \hat{\phi}(2\ell)$, so that $\mathbf{b}\in \mathcal{B}(s)$. We note that
\begin{eqnarray*}
\sigma_{2^n}(\mu_q^*;f,\x)&=&\int_{\SS^q} \phi(\x\cdot\y)\mathcal{D}_\phi(\sigma_{2^n}(\mu_q^*;f))(\y)d\mu_q^*(\y)\\
&=& \int_{\SS^q} \phi(\x\cdot\y)\sigma_{2^n}(\mu_q^*;\mathcal{D}_\phi(f),\y)d\mu_q^*(\y)\\
&=& \int_{\SS^q} \sigma_{2^n}(\mu_q^*;\phi(\x\cdot\circ),\y)\mathcal{D}_\phi(f)(\y)d\mu_q^*(\y)\\
&=& \int_{\SS^q}\tilde{\Phi}_{2^n}(h, \mathbf{b}; \x\cdot\y)\mathcal{D}_\phi(f)(\y)d\mu_q^*(\y).
\end{eqnarray*}
Therefore,
$$
f(\x)-\sigma_{2^n}(\mu_q^*;f,\x)=\int_{\SS^q}\left\{\phi(\x\cdot\y)-\tilde{\Phi}_{2^n}(h, \mathbf{b}; \x\cdot\y)\right\}\mathcal{D}_\phi(f)(\y)d\mu_q^*(\y)=U_{2^n}(\mu_q^*;\mathcal{D}_\phi(f)).
$$
We now deduce \eref{mustardegapprox} using Proposition~\ref{fundaprop} with $\mu_q^*$ in place of $\nu$ and $2^{-n}$ in place of $d$.

Specializing to the case $p=\infty$, Proposition~\ref{sigmaopprop} shows that
$$
\|f-\sigma_{2^n}(\mu;f)\|_\infty \le c\tn\mu\tn_{R, 2^{-n}}E_{2^n,\infty}(f)\le c\tn\mu\tn_{R, 2^{-n}}\|f-\sigma_{2^{n-2}}(\mu_q^*;f)\|_\infty.
$$
Therefore, \eref{mustardegapprox} implies \eref{mudegapprox}.
\end{Proof}

\noindent\textsc{Proof of Theorem~\ref{mainshallowtheo}.}\\
We use Lemma~\ref{polyapproxlemma} with $P=\sigma_{2^n}(\mu_q^*;f)$ in place of $P$, and recall that
$$
\|P\|_p=\|\mathcal{D}_\phi(\sigma_{2^n}(\mu_q^*;f))\|_p=\|\sigma_{2^n}(\mu_q^*;\mathcal{D}_\phi(f))\|_p\le c\|\mathcal{D}_\phi(f)\|_p
$$
to conclude that
\bea\label{pf4eqn2}
\|\sigma_{2^n}(\mu_q^*;f)-G_{2^n}(\phi,\mu^*,\nu;f)\|_p&=&\left\|P-\int_{\SS^q}\phi(\circ\cdot\y)\mathcal{D}_\phi(P)(\y)d\nu(\y)\right\|_p\nonumber\\
& \le& 
 c2^{-n(s-(q-1)/2)}\tn\nu\tn_{R,1/2^n}^{1/p'}\|\mathcal{D}_\phi(P)\|_{\nu;p}
\eea
In view of Proposition~\ref{mzequivprop}(c),
$$
\|\mathcal{D}_\phi(P)\|_{\nu;p}\le c\tn\nu\tn_{R,1/2^n}^{1/p}\|\mathcal{D}_\phi(P)\|_p.
$$
Therefore, 
the estimate \eref{shallowdegapprox} follows from \eref{mustardegapprox} and \eref{pf4eqn2}. The estimate \eref{sphcoefftheory} is a simple consequence of 
Proposition~\ref{sigmaopprop}.

The estimate \eref{sampleshallowdegapprox} is proved in the same way, using 
\eref{mudegapprox} instead of \eref{mustardegapprox}.
\qed

\begin{appendix}
\renewcommand{\theequation}{\Alph{section}.\arabic{equation}}

\bhag{Appendix: Computation of coefficients}\label{coeffsect}

The main purpose of this appendix is to prove the following proposition.
\begin{prop}\label{actcoeffprop}
Let $\alpha>-1$, $\gamma >-1/2$ and $2\gamma+1$ not be an even integer. Then for $\ell =0,1,\cdots$,
\bea\label{actcoeff1}
\int_{-1}^1 |t|^{2\gamma+1}p_{2\ell }^{(\alpha,\alpha)}(t)(1-t^2)^\alpha dt &=&
(-1)^\ell \frac{\cos(\pi\gamma)\alpha!(2\gamma+1)!}{2^{2\gamma+1}\sqrt{\pi}} \frac{(\ell -\gamma-3/2)!}{(\ell+\gamma +\alpha+1)!}p_{2\ell }^{(\alpha,\alpha)}(1)\nonumber\\
&=& \cos(\pi\gamma)\frac{(2\gamma+1)!}{2^{2\gamma+1}}\frac{(\ell +\alpha)!}{(\ell -1/2)!}\frac{(\ell -\gamma-3/2)!}{(\ell+\gamma +\alpha+1)!}p_{2\ell }^{(\alpha,\alpha)}(0).
\eea
In particular, the following asymptotic expansions hold with  real constants $c_{1,j}, c_{2,j}$:
\be\label{actcoeff2}
\int_{-1}^1 |t|^{2\gamma+1}p_{2\ell }^{(\alpha,\alpha)}(t)(1-t^2)^\alpha dt=(-1)^\ell \frac{p_{2\ell }^{(\alpha,\alpha)}(1)}{\ell^{2\gamma+2+\alpha+1/2}}\sum_{j=0}^\infty \frac{c_{1,j}(\gamma,\alpha)}{\ell^j}=\frac{p_{2\ell }^{(\alpha,\alpha)}(0)}{\ell^{2\gamma+2}}\sum_{j=0}^\infty \frac{c_{2,j}(\gamma,\alpha)}{\ell^j}.
\ee
\end{prop}

In preparation for the proof, we make first some observations. First, we note that the duplication and complimentary formulas for gamma functions take the form
\be\label{gammaduplicate}
\frac{(2z)!}{z!}=\frac{2^{2z}(z-1/2)!}{\sqrt{\pi}},
\ee
and
\be\label{gammacomp}
\frac{1}{(-z)!}=\frac{\sin (\pi z)}{\pi z}z!= \frac{\sin (\pi z)}{\pi }(z-1)!, \qquad z \mbox{ {\rm not a negative integer}},
\ee
respectively.

In view of  \eref{rodrigues} and \eref{pkat1}  we obtain
\be\label{pkoverpk1}
(1-x^2)^\alpha p_{2\ell}^{(\alpha,\alpha)}(x) = p_{2\ell}^{(\alpha,\alpha)}(1)\frac{\alpha!}{2^{2\ell}(2\ell+\alpha)!}\frac{d^{2\ell}}{dx^{2\ell}}(1-x^2)^{2\ell+\alpha}.
\ee
Further,  since
$$
(1-x^2)^{2\ell+\alpha}=\sum_{j=0}^\infty \frac{(2\ell+\alpha)!}{j! (2\ell+\alpha-j)!}(-1)^j x^{2j}, \qquad |x|<1,
$$
 we
deduce that for integer $m\ge 0$,
\be\label{wtderivative}
\frac{d^{2m}}{dx^{2m}}(1-x^2)^{2\ell+\alpha}\bigg|_{x=0}=(-1)^m\frac{(2\ell+\alpha)!(2m)!}{m! (2\ell+\alpha-m)!}.
\ee
In particular, \eref{rodrigues} shows that
$$
p_{2\ell}^{(\alpha,\alpha)}(0)=(-1)^\ell \frac{(2\ell+\alpha)!}{2^{2\ell}\ell!(\ell+\alpha)!}
\left\{\frac{4\ell+\alpha+\beta+1}{2^{\alpha+\beta+1}}\frac{(2\ell)!(2\ell+\alpha+\beta)!}{(2\ell+\alpha)!(2\ell+\beta)!}\right\}^{1/2}.
$$
Hence,
 using  \eref{gammaduplicate} and \eref{pkat1}, we see that
\be\label{pkat1over0}
\frac{p_{2\ell}^{(\alpha,\alpha)}(1)}{p_{2\ell}^{(\alpha,\alpha)}(0)}=(-1)^\ell \frac{2^{2\ell}\ell!}{(2\ell)!}\frac{(\ell+\alpha)!}{\alpha!}=\frac{(-1)^\ell\sqrt{\pi}}{\alpha!}\frac{(\ell+\alpha)!}{(\ell-1/2)!}.
\ee

\noindent\textsc{Proof  of Proposition~\ref{actcoeffprop}.}
In this proof, let
\be\label{igammak}
I_{\gamma,\ell }=\int_0^1 x^{2\gamma+1}\frac{d^{2\ell }}{dx^{2\ell }}(1-x^2)^{2\ell +\alpha}dx.
\ee
We will prove that
\bea\label{igammakcompute}
I_{\gamma,\ell }&=& (-1)^{\ell } \cos(\pi\gamma)\frac{(2\gamma+1)! (2\ell -2\gamma-2)!(2\ell +\alpha)!}{(\ell -\gamma-1)!(\gamma+\ell +\alpha+1)!}\nonumber\\
&=&(-1)^\ell\frac{\cos(\pi\gamma)(2\gamma+1)!}{2^{2\gamma+2}\sqrt{\pi}}\frac{2^{2\ell}(2\ell+\alpha)!(\ell-\gamma-3/2)!}{(\ell+\gamma+\alpha+1)!}.
\eea

The second equation in \eref{igammakcompute} follows from the first using \eref{gammaduplicate}. So, we need to prove only the first equation.
We distinguish two cases.

\noindent\textbf{Case I}: $2\ell \le 2\gamma+1$, or  $2\gamma+1$ is not an integer.\\
Integration by parts $2\ell $ times  in \eref{igammak} gives
\bea\label{caseIcompute1}
I_{\gamma,\ell }&=&\frac{(2\gamma+1)!}{(2\gamma+1-2\ell )!}\int_0^1 x^{2\gamma+1-2\ell }(1-x^2)^{2\ell +\alpha}dx= \frac{(2\gamma+1)!}{2(2\gamma+1-2\ell )!}\int_0^1 y^{\gamma-\ell }(1-y)^{2\ell +\alpha}dy\nonumber\\
& =&\frac{(2\gamma+1)!(\gamma-\ell )! (2\ell +\alpha)!}{2(2\gamma+1-2\ell )!(\gamma+\ell +\alpha+1)!},
\eea
where we note that neither $\gamma-\ell $ nor $2\gamma+1-2\ell $ is a negative integer.
Using \eref{gammacomp}, 
$$
\frac{(\gamma-\ell )!}{(2\gamma+1-2\ell )!}=-\frac{(2\ell -2\gamma-2)!\sin((2\gamma+1)\pi)}{(\ell -\gamma-1)!\sin(\pi (\ell -\gamma))}=2(-1)^\ell  \cos(\pi\gamma)
\frac{(2\ell -2\gamma-2)!}{(\ell -\gamma-1)!}.
$$
Together with \eref{caseIcompute1}, this leads to \eref{igammakcompute}.\\

\noindent\textbf{Case II:} $2\gamma+1$ is an odd integer (i.e., $\gamma$ is an integer), and $2\ell > 2\gamma+1$.\\
Let $m=\ell -\gamma-1$. Then $m\ge 0$ is an integer. Using Leibnitz rule, we find that
$$
\frac{d^{2m}}{dx^{2m}}(1-x^2)^{2\ell +\alpha}\bigg|_{x=1}=0.
$$
Therefore, an integration by parts  $2\gamma+1$ times gives (cf. \eref{wtderivative})
\bea\label{caseIIIcompute1}
I_{\gamma,\ell }&=&(-1)^{2\gamma+1}(2\gamma+1)! \int_0^1 \frac{d^{2\ell -2\gamma-1}}{dx^{2\ell -2\gamma-1}}(1-x^2)^{2\ell +\alpha}dx=(-1)^{\ell +\gamma}\frac{(2\gamma+1)!(2\ell +\alpha)!(2m)!}{m! (2\ell +\alpha-m)!}\nonumber\\
&=& (-1)^{\ell +\gamma}\frac{(2\gamma+1)!(2\ell +\alpha)!(2\ell -2\gamma-2)!}{(\ell -\gamma-1)! (\ell +\alpha+\gamma+1)!}\nonumber\\
&=& (-1)^{\ell } \cos(\pi\gamma)\frac{(2\gamma+1)! (2\ell -2\gamma-2)!(2\ell +\alpha)!}{(\ell -\gamma-1)!(\gamma+\ell +\alpha+1)!}.
\eea
This completes the proof of \eref{igammakcompute} also in this case.

Having proved \eref{igammakcompute}, the first identity in \eref{actcoeff1}
now follows from \eref{pkoverpk1}. The second identity in \eref{actcoeff1} follows from the first and \eref{pkat1over0}. 
The expansions \eref{actcoeff2} follow from \eref{actcoeff1} and the asymptotic formula for the ratios of gamma functions \cite[Chapter~4, formula~(5.02)]{olverbook}. 
\qed
\end{appendix}


\end{document}